\documentclass[]{SPAICE}
\usepackage{multirow}
\usepackage{pifont}
\usepackage{subcaption}
\usepackage{siunitx}
\usepackage{authblk}

\def\AuthorShort{M. Chen et al.}

\author[1]{Maggie Chen\thanks{These authors contributed equally and are listed in alphabetical order.}}
\author[1]{Hala Lamdouar\protect\footnotemark[1]}
\author[2]{Luca Marini\protect\footnotemark[1]}
\author[3]{Laura Martínez-Ferrer}
\author[4]{Chris Bridges}
\author[5]{Giacomo Acciarini}
\affil[1]{University of Oxford, Oxford, United Kingdom}
\affil[2]{Delft University of Technology, Delft, The Netherlands}
\affil[3]{Universitat de Val\`encia, Val\`encia, Spain}
\affil[4]{University of Surrey, Guildford, United Kingdom}
\affil[5]{European Space Agency, Advanced Concepts Team, Noordwijk, The Netherlands}

\title{Methane Detection On Board Satellites from Unorthorectified Imagery}

\begin{document}

\maketitle

\begin{abstract}
As a potent greenhouse gas, methane is a major driver of climate change. Its effective mitigation relies on timely detection. Conventional detection methods rely on \textit{orthorectification} to correct geometric distortions and \textit{matched filters} to enhance plume signals, which are  steps designed for ground processing and poorly suited to onboard execution.
We introduce UnorthoDOS, a dataset and approach for training machine learning models directly on unorthorectified hyperspectral imagery, bypassing both orthorectification and matched-filter products. 
Our U-Net models trained on unorthorectified data approach the performance of models trained on orthorectified data (IoU 16.91\% vs. 18.47\% on all plumes), while both substantially outperform the mag1c matched-filter baseline (IoU 4.76\%).
We further demonstrate the feasibility of onboard deployment: FP16 compression halves model size with under 0.3\% output deviation. 
The trained ML models and two ML-ready datasets -- orthorectified and unorthorectified hyperspectral imagery from the EMIT sensor -- are publicly available at \url{https://huggingface.co/datasets/SpaceML/UnorthoDOS}, with code at \url{https://github.com/spaceml-org/plume-hunter}.
\end{abstract}

\section{Introduction}
Methane has a global warming potential approximately 84 times greater than that of carbon dioxide over a 20-year period \cite{myhre2014}. A significant proportion of point-source methane emissions originates from “super-emitters” in the oil and gas sector, which are detectable from space \cite{jacob2016satellite}. 
As interest grows in deploying on board artificial intelligence (AI) for Earth observation (EO), real‑time processing on satellites is becoming technically and operationally feasible \cite{11474746, rijlaarsdam2024next, furano2020towards, giuffrida2021varphi, marin2021phi}. When coupled with hyperspectral imaging, whose fine spectral resolution is well suited to identifying methane, on board AI provides a unique opportunity for rapid methane detection, reducing latency and downlink requirements \cite{11474746, ghasemi2025onboard, marini2025semi}. This combination offers a pathway to faster response and more effective mitigation of high-impact methane releases.
Methane plume detection in hyperspectral satellite imagery typically involves two key processing steps: (1) \textit{orthorectification}, which corrects geometric distortions caused by sensor viewing angle, terrain variations, and Earth curvature, and (2) the generation of methane enhancement products, often using \textit{matched filters} \cite{manolakis2013detection, thompson2015real, foote2020fast} to enhance weak plume signals by comparing against predefined spectral signatures. 
However, orthorectification is intended for ground processing and not designed on board execution \cite{11474746, meoni2024unlocking}, and traditional matched filters are iterative algorithms with high computational cost and are susceptible to high false-positive rates \cite{herec2025optimizing}. Previous studies such as \cite{STARCOP1} improved segmentation accuracy by combining matched filter outputs with Red Green Blue (RGB) images using Deep Learning models. More recently, end-to-end approaches 
used lightweight Vision Transformers \cite{ruuvzivcka2025hyperspectralvits} to advance performance, and efficient, low-power
algorithms have been proposed in \cite{herec2025optimizing}. Summarised in \cref{table:datasets}, these methods remain reliant on orthorectified data. This study introduces UnorthoDOS, an unorthorectified hyperspectral dataset and approach that better reflects on board acquisitions, and shows that methane plume detection trained directly on unorthorectified imagery achieves performance similar to that of orthorectified pipelines. It extends \cite{chen2025towards} with onboard model compression results, increasing the feasibility of deployment on resource-constrained hardware.

\begin{table*}[tbh!]
\begin{center}
\small
\begin{tabular}{lcrcc}
\toprule
\textbf{Dataset} & \textbf{Instrument} & \textbf{Bands} & \textbf{Spectral range (nm)} & \textbf{Unorthorectified} \\
\midrule
STARCOP \cite{STARCOP1} & AVIRIS-NG & 125 & RGB, 1573--1699, 2004--2480 & \ding{55}\\
\cmidrule(lr){2-4}
OxHyperSyntheticCH4 \cite{ruuvzivcka2025hyperspectralvits} & EMIT & 86 & RGB, 1573--1699, 2004--2478 & \ding{55} \\
OxHyperRealCH4 \cite{ruuvzivcka2025hyperspectralvits} & EMIT & 86 & RGB, 1573--1699, 2004--2478 & \ding{55} \\
UnorthoDOS (Ours) & EMIT & 86 & RGB, 1573--1699, 2004--2478 & \ding{51} \\
\bottomrule
\end{tabular}
\caption{Comparison of hyperspectral datasets for methane plume detection, with emphasis on the presence or absence of orthorectification preprocessing.}
\label{table:datasets}
\end{center}
\end{table*}

\section{Methodology}


\subsection{Hyperspectral \& Methane Data Source}
Datasets are constructed from hyperspectral observations acquired by the EMIT imaging spectrometer aboard the International Space Station (ISS), obtained via the NASA Earthdata portal\footnote{\url{https://search.earthdata.nasa.gov/}}. The L1B at-sensor radiance data product \cite{L1BEMIT} provides 285 hyperspectral bands at 60 m spatial resolution, and the L2B methane plume complex product \cite{Methane_plumes}, also at 60 m resolution, supplies orthorectified ground-truth plume masks. 
L2B annotations are available only in orthorectified form, 
which motivates the unorthorectification procedure described in \cref{sec:unorthodos_generation}. The full corpus spans 1,574 annotated methane plumes drawn from EMIT scenes acquired between August 10, 2022 and October 26, 2024. %
Following the band-selection procedure of \cite{ruuvzivcka2025hyperspectralvits}, 86 of the 285 bands are retained, spanning $1573\text{-}1699$ nm and $2004\text{-}2478$ nm. This subset covers the primary methane absorption features while including a small number of non-absorption bands and three RGB bands to let the model learn background spectral variability.


\subsection{UnorthoDOS Dataset Generation}
\label{sec:unorthodos_generation}

\begin{figure*}[]
    \centering
    \includegraphics[width=.95\textwidth]{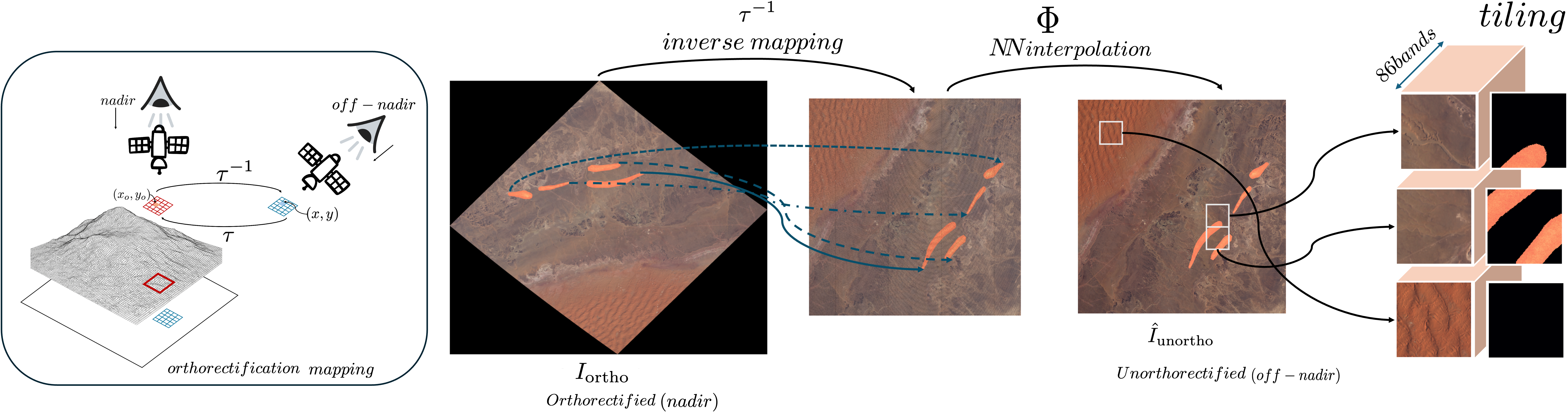}
    \caption{Overview of the UnorthoDOS pipeline for synthesizing unorthorectified hyperspectral data from orthorectified EMIT data.}
    \label{fig:ortho_transfomation}
\end{figure*}

The method to generate unorthorectified data is illustrated schematically in \cref{fig:ortho_transfomation}.
Orthorectification is the mapping $\tau$ that transforms pixel coordinates $(x, y)$ from an angled (\textit{off-nadir}) image $I_{\text{unortho}}$ to coordinates $(x_o, y_o)$ in the orthogonal (\textit{nadir}) image $I_{\text{ortho}}$, compensating for sensor viewing geometry and terrain relief. This mapping is extracted directly from the geometric lookup table included in the EMIT L1B product. 
\begin{align}
\tau  : \begin{array}{rcl}
I_{\text{unortho}} & \longrightarrow & I_{\text{ortho}}\\
(x,y) & \longmapsto     & \tau(x,y) = (x_o,y_o)
\end{array}
\end{align}

Unorthorectification is the inverse mapping ${\tau}^{-1}$, approximated by reconstructing $I_{\text{unortho}}$ from $I_{\text{ortho}}$ using a pixel coordinate grid in the source (unorthorectified) plane. The grid is then transformed forward through $\tau$ into the orthorectified plane, and used to sample pixels from the orthorectified methane plume annotations back into the unorthorectified plane.

Because ${\tau}$ is not bijective, $I_{\text{unortho}}$ and $I_{\text{ortho}}$ may differ in size. Consequently, the inverse mapping ${\tau}^{-1}$ is not necessarily surjective, leading to some pixels in $I_{\text{unortho}}$ without corresponding values in $I_{\text{ortho}}$.
These missing values in $I_{\text{unortho}}$ are filled using nearest-neighbor interpolation $\Phi$. In summary, the procedure to generate an approximate \textit{off-nadir} image can be expressed as:
\begin{equation}
\hat{I}_{\text{unortho}} = \Phi \circ {\tau}^{-1} (I_{\text{ortho}}) \text{.}
\end{equation}
Both hyperspectral images and their corresponding plume annotations are tiled into $128\times128$ pixel tiles for model training. 
The resulting unorthorectified ML-ready dataset is referred to as \textbf{UnorthoDOS} (\textbf{Unortho}rectified \textbf{D}ataset for \textbf{O}nboard \textbf{S}atellite methane detection) \cite{chen2025unorthodos}.


\subsection{Orthorectified Benchmark Dataset}
To isolate the effect of orthorectification on model performance, the orthorectified benchmark dataset is built from the same underlying EMIT scenes and 1,574 plumes as UnorthoDOS, using the orthorectified L1B radiance images paired with the original orthorectified L2B plume annotations and applying the tiling procedure described in \cref{sec:unorthodos_generation}. 

\subsection{Experimental Setup}
\label{sec:exp_stup}
\begin{figure}
    \centering
    \includegraphics[width=\columnwidth,trim=0 48 0 50, clip]{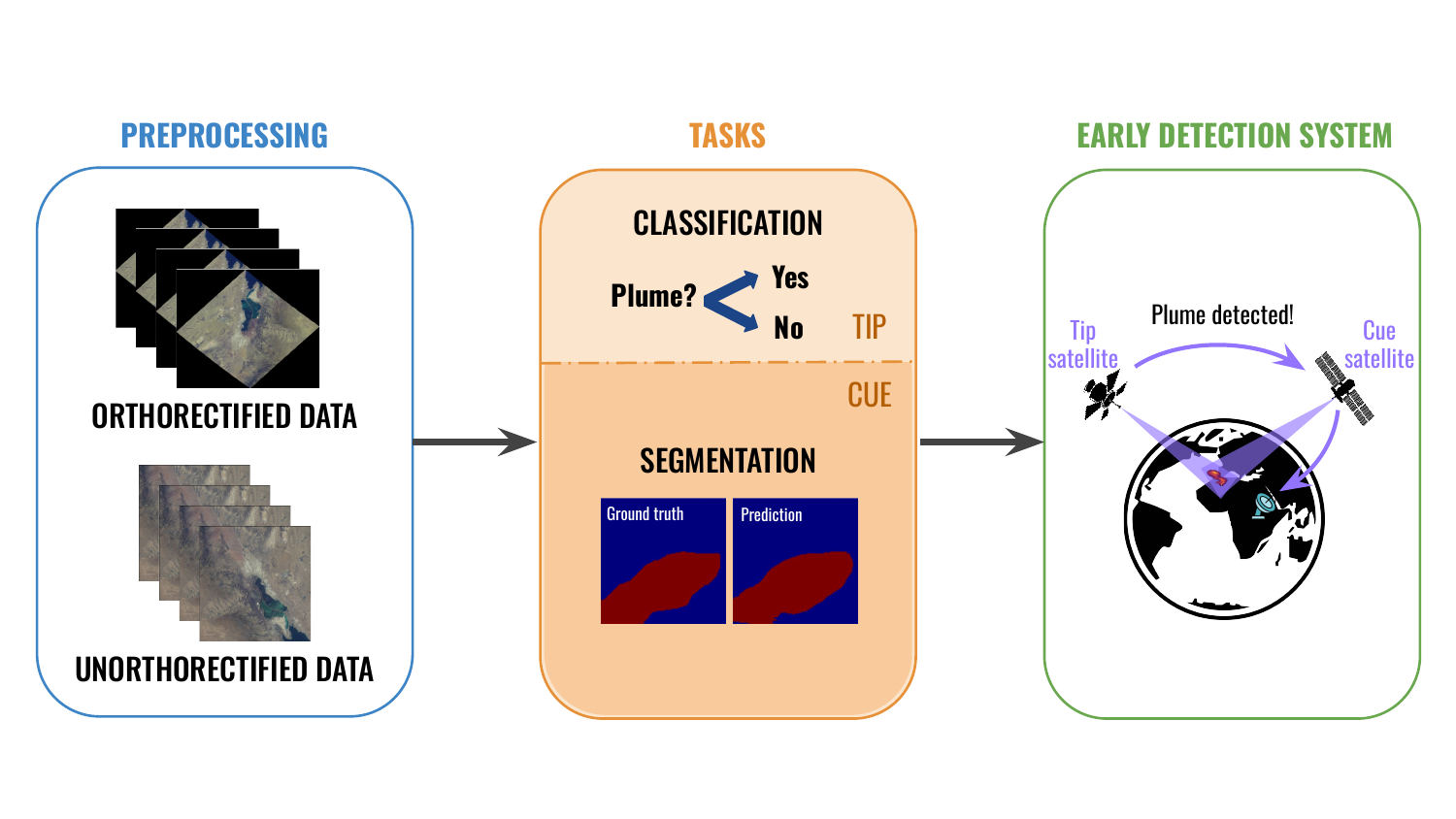}
    \caption{Overview of the detection pipeline. EMIT images are preprocessed into orthorectified and unorthorectified datasets for ML model training.
    Trained ML models simulate a tip and cue setting to perform plume classification and segmentation.}
    \label{fig:workflow}
\end{figure}

To support a rapid detect-and-response pipeline, we adopt a simplified \textit{tip and cue} paradigm \cite{ceos} (\cref{fig:workflow}).
A tip satellite performs lightweight binary classification to flag candidate methane plumes, and a cue satellite performs semantic segmentation to precisely localize them.

Data splits are performed at the scene (L1B image) level, prior to tiling, to avoid tile leakage. Each of the orthorectified and unorthorectified datasets is split into a training set (80\% of scenes), a validation set (15\% of scenes), and a holdout test set (5\% of scenes), with identical scene-level splits used across the two geometric variants. Given the limited number of annotated methane plumes, training and validation sets are augmented with random spatial translations. Input images values are normalised per spectral band by subtracting the training-set mean and dividing by the training-set standard deviation.

A U-Net \cite{ronneberger2015u}, adapted to accept 86-channel hyperspectral input, is used to predict binary per-pixel methane plume masks\footnote{Full architecture and augmentation configuration are provided in the released code.}.
A tile is classified as plume-positive if its predicted mask contains at least one positive pixel.
Class imbalance between plume and background is addressed with the Dice loss \cite{sudre2017generalised}:
\begin{equation}
    L_{Dice} = 1 - D = 1 - \frac{2 TP}{2 TP + FP + FN}\text{ .}
\end{equation}

Two U-Net models are trained separately on the orthorectified and unorthorectified datasets with Adam \cite{kingma2015adam}, learning rate $10^{-4}$, batch size 32, for 100 epochs, selecting the best model with lowest Dice validation loss. Training is performed on a single 40 GB NVIDIA A100 GPU.

\subsection{Performance Evaluation} 
The trained PyTorch models (prior to compression) are evaluated on the holdout test sets described in \cref{sec:exp_stup}. 
We report precision, recall, and F1-score for both classification and segmentation, classification accuracy, and segmentation IoU, all computed at a 0.5 sigmoid threshold.
To assess performance as a function of plume strength, we additionally report all metrics on a subset restricted to tiles whose maximum annotated methane concentration is $\geq$ 900 ppm m.

\subsection{Model Compression}
Trained U-Net checkpoints are exported from PyTorch to the Open Neural Network Exchange (ONNX) format~\cite{onnxruntime}. 
First, a ONNX graph optimization compresses models to 32-bit floating-point model (FP32) that is numerically equivalent to the original PyTorch checkpoint. This model is then further compressed to 16-bit floating-point precision (FP16), and separately to 8-bit integer precision (INT8) using weight-only quantisation.
For each precision level, we report file size and the maximum per-pixel relative difference between compressed and original model outputs, computed on an NVIDIA RTX 4070 Ti SUPER GPU (16 GB).

\section{Results}
\begin{figure*}[htp!]
\begin{subfigure}{0.475\textwidth}
    \centering
    \parbox[c]{0.32\textwidth}{\centering RGB \& \\ ground truth}%
    \makebox[0.32\textwidth]{U-Net}%
    \makebox[0.32\textwidth]{Mag1c}\\
    \begin{minipage}[t]{0.32\textwidth}
        \includegraphics[width=\textwidth]{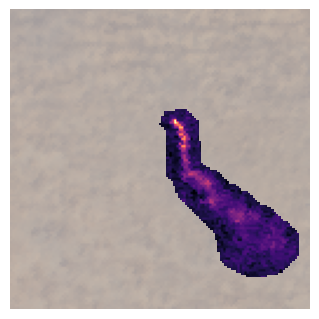}
        \includegraphics[width=\textwidth]{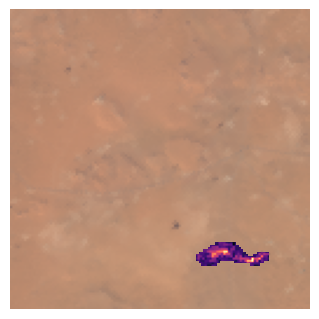}
    \end{minipage}
    \begin{minipage}[t]{0.32\textwidth}
        \includegraphics[width=\textwidth]{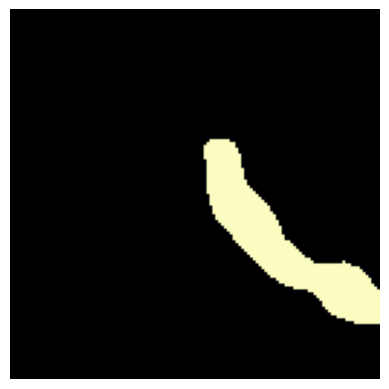}
        \includegraphics[width=\textwidth]{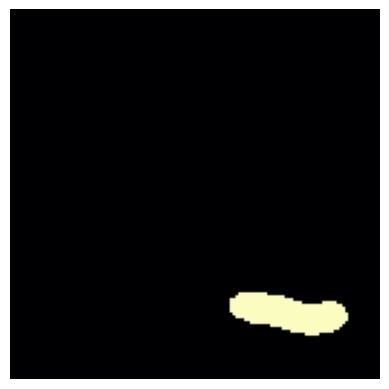}
    \end{minipage}
    \begin{minipage}[t]{0.32\textwidth}
        \includegraphics[width=\textwidth]{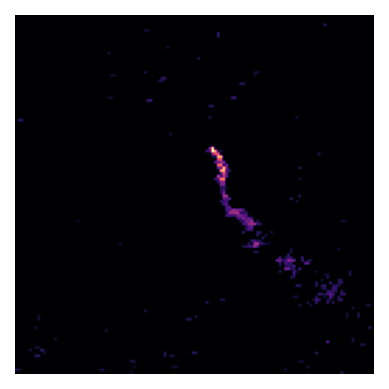}
        \includegraphics[width=\textwidth]{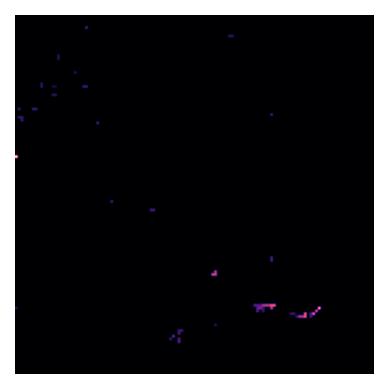}
    \end{minipage}
    \label{fig:qualitative_unortho_fig}
\caption{Orthorectified}
\label{fig:ortho}
\end{subfigure}%
\hspace{0.04\textwidth}
\begin{subfigure}{0.475\textwidth}
    \centering
    \parbox[c]{0.32\textwidth}{\centering RGB \& \\ ground truth}%
    \makebox[0.32\textwidth]{U-Net}%
    \makebox[0.32\textwidth]{Mag1c}\\
    \begin{minipage}[t]{0.32\textwidth}
        \includegraphics[width=\textwidth]{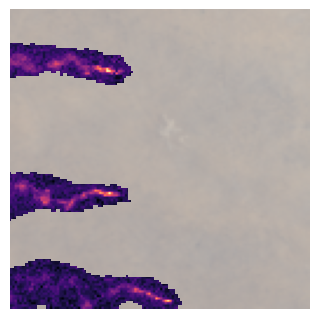}
        \includegraphics[width=\textwidth]{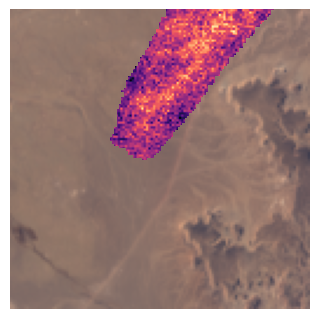}
    \end{minipage}
    \begin{minipage}[t]{0.32\textwidth}
        \includegraphics[width=\textwidth]{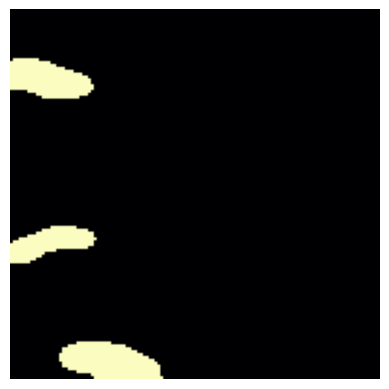}
        \includegraphics[width=\textwidth]{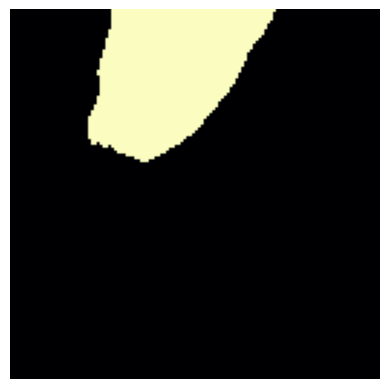}
    \end{minipage}
    \begin{minipage}[t]{0.32\textwidth}
        \includegraphics[width=\textwidth]{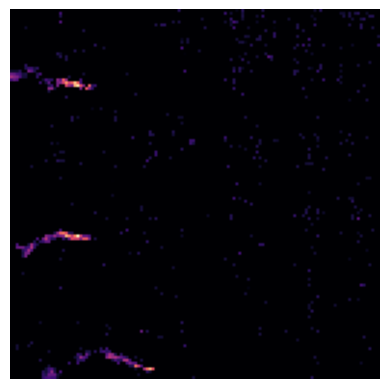}
        \includegraphics[width=\textwidth]{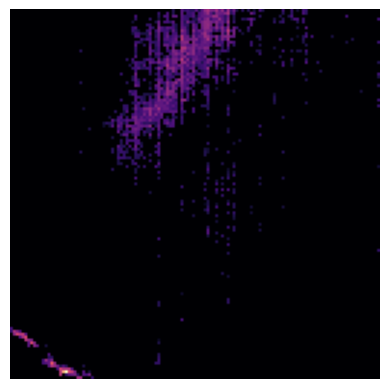}
    \end{minipage}
\caption{Unorthorectified}
\label{fig:unortho}
\end{subfigure}
\caption{Visualisation of semantic segmentation results on 2 example tiles from the orthorectified (\cref{fig:ortho}) and the unorthorectified (\cref{fig:unortho}) datasets each. From \textit{Left} to \textit{Right} in each sub-figure: L1B tiles (only RGB bands shown for visualisation) overlaid with ground truth methane plume annotations; predicted semantic segmentation plume masks from U-Net; segmentation predictions from mag1c~\cite{foote2020fast}.}
\label{fig:qualitative_unortho}
\end{figure*}

\cref{table:results_classification,table:results_segmentation} report image classification and semantic segmentation performance for the U-Net models and the mag1c~\cite{foote2020fast} baseline.
For semantic segmentation, the orthorectified U-Net improves upon mag1c by 13.71 IoU points on all plumes (18.47\% vs. 4.76\%), and by 25.96 points on strong plumes (30.80\% vs. 4.84\%); the unorthorectified U-Net shows a comparable margin (16.91\% vs. 4.76\% and 26.17\% vs. 4.84\%, respectively). 
Qualitatively (\cref{fig:qualitative_unortho}), U-Net predictions are visually coherent and closely follow plume boundaries, whereas mag1c produces fragmented outputs and a higher false-positive rate from misidentified terrestrial features, visible in the second unorthorectified example. 

Both segmentation and classification improve substantially when restricted to strong plumes, for both data regimes (\cref{table:results_classification,table:results_segmentation}). Segmentation IoU rises from 18.47\% to 30.80\% (orthorectified) and from 16.91\% to 26.17\% (unorthorectified), while classification recall rises from 56.21\% to 83.33\% and from 48.89\% to 71.61\%, respectively. This pattern indicates that model performance is limited primarily by weak, low-concentration plumes rather than by a general failure to localize methane signal, a limitation we return to in \cref{sec:discussion}.

U-Net models trained on the unorthorectified and orthorectified data achieve comparable performances: IoU on all plumes is 16.91\% vs. 18.47\%, and on strong plumes 26.17\% vs. 30.80\%. Classification performance shows a wider gap: accuracy is 5.3 points lower on all plumes (71.48\% vs. 76.80\%) and 6.8 points lower on strong plumes (85.65\% vs. 92.41\%) for the unorthorectified case (\cref{table:results_classification}). 
Despite this gap, both settings substantially outperform the mag1c baseline, and the unorthorectified model requires no orthorectification step at inference time. This constitutes, to our knowledge, the first demonstration that methane plume detection is feasible directly on unorthorectified satellite imagery, without the geometric correction step conventionally treated as a prerequisite.

\begin{table*}[tbh!]
\begin{center}
\small
\begin{tabular}{llccccc}
\toprule
Setting & Model & Threshold (ppm m) & Precision & Recall & F1-Score & Accuracy (\%) \\
\midrule
\multirow{2}{*}{Orthorectified} & \multirow{2}{*}{mag1c \cite{foote2020fast}} & N/A & 52.55 & \textbf{94.12} & 67.45 & 54.58  \\
 & & $\geq$ 900 & 39.60 & 95.24 & 55.94 & 46.84  \\
\cmidrule(lr){1-7}
\multirow{2}{*}{Orthorectified} & \multirow{2}{*}{U-Net} & N/A & \textbf{95.56} & 56.21 & 70.78 & 76.80  \\
 & & $\geq$ 900 & 94.60 & 83.33 & \textbf{88.61} & \textbf{92.41}  \\
\cmidrule(lr){1-7}
\multirow{2}{*}{Unorthorectified} & \multirow{2}{*}{U-Net} & N/A & 89.19 & 48.89 & 63.16 & 71.48  \\
 & & $\geq$ 900 & 87.88 & 71.61 & 78.91 & 85.65  \\
\bottomrule
\end{tabular}
\caption{Image classification performance. Precision, recall, F1-score, and accuracy are reported for mag1c \cite{foote2020fast} and U-Net, trained/evaluated on orthorectified and unorthorectified data. Best result per column within each block is highlighted in \textbf{bold}.}
\label{table:results_classification}
\end{center}
\end{table*}

\begin{table*}[tbh!]
\begin{center}
\small
\begin{tabular}{llccccc}
\toprule
Setting & Model & Threshold (ppm m) & Precision & Recall & F1-Score & IoU \\
\midrule
\multirow{2}{*}{Orthorectified} & \multirow{2}{*}{mag1c \cite{foote2020fast}} & N/A & 41.67 & 15.69 & 22.80 & 4.76 \\
 & & $\geq$ 900 & 44.91 & 15.69 & 23.25 & 4.84 \\
\cmidrule(lr){1-7}
\multirow{2}{*}{Orthorectified} & \multirow{2}{*}{U-Net} & N/A & 79.23 & 19.91 & 31.82 & 18.47 \\
 & & $\geq$ 900 & 82.77 & 32.52 & 46.69 & \textbf{30.80} \\
\cmidrule(lr){1-7}
\multirow{2}{*}{Unorthorectified} & \multirow{2}{*}{U-Net} & N/A & 88.41 & 19.85 & 32.42 & 16.91 \\
 & & $\geq$ 900 & \textbf{89.08} & \textbf{33.01} & \textbf{48.16} & 26.17 \\
\bottomrule
\end{tabular}
\caption{Semantic segmentation performance. Precision, recall, F1-score, and IoU are reported for mag1c \cite{foote2020fast} and U-Net, trained/evaluated on orthorectified and unorthorectified data. Best result per column within each block is highlighted in \textbf{bold}.}
\label{table:results_segmentation}
\end{center}
\end{table*}

Model compression substantially reduces file size (\cref{table:compression}): FP16 conversion halves the model size (3.30 to 1.70 MB) with a maximum per-pixel output deviation below 0.3\%, while INT8 quantisation reduces size further (to 0.84 MB, a 4$\times$ reduction) at the cost of up to $\sim$12\% maximum output deviation. We consider FP16 the more robust operating point for onboard deployment given its negligible impact on outputs, while INT8's larger deviation warrants further validation before deployment on hardware with stricter memory budgets.


\begin{table*}[htp!]
\begin{center}
\small
\resizebox{0.788\linewidth}{!}{%
\begin{tabular}{llcccc}
\toprule
& & \textbf{PyTorch} & \textbf{32-bit ONNX} & \textbf{16-bit ONNX} & \textbf{8-bit Quant} \\
\midrule
 \textbf{File size (MB)} & & 3.30 & 3.25 & 1.70 & 0.84 \\
 \midrule
 \multirow{2}{*}{\textbf{Max relative difference (\%)}} & Orthorectified & 0 & 1.66$\times 10^{-4}$ & 0.30 & 11.93 \\
   & Unorthorectified & 0 & 2.19$\times 10^{-4}$ & 0.25 & 10.70 \\
\bottomrule
\end{tabular}
}
\end{center}
\caption{File sizes in MB, 
and the maximum relative difference in the outputs of the original PyTorch models trained on the orthorectifed and unorthorectified datasets and ONNX models compressed to varying precisions.
}
\label{table:compression}
\end{table*}

\section{Discussion}
\label{sec:discussion}
In this work, we present UnorthoDOS, 
a dataset and training approach for methane plume detection directly on unorthorectified satellite imagery. 
We show that models trained on orthorectified and unorthorectified data achieve comparable performance, demonstrating that orthorectification can be bypassed: a critical advantage for real-time detection on resource-constrained satellites. We further demonstrate onboard deployment feasibility via compression, halving model size with <1\% output deviation. A key limitation is reduced sensitivity to weak methane plumes, mitigable by training on larger datasets as hyperspectral satellite deployments are increasing.
\begin{acknowledgments}
    This work has been enabled by Frontier Development Lab Earth Systems Lab (https://eslab.ai/) a public/private partnership between the European Space Agency (ESA), Trillium Technologies, the University of Oxford and leaders in commercial AI supported by Google Cloud, Scan Computers, Nvidia Corporation and Pasteur Labs. 
\end{acknowledgments}
\printbibliography
\addcontentsline{toc}{section}{References}

\end{document}